\documentclass[10pt,journal,compsoc]{IEEEtran}
\usepackage{comment}
\usepackage[ruled,vlined,linesnumbered]{algorithm2e}
\usepackage{amssymb,graphicx,amsmath}
\SetKwComment{Comment}{ $\triangleright$ \ }{}
\usepackage{booktabs}
\usepackage{multirow}
\usepackage{subfigure}
\usepackage{color}
\usepackage[normalem]{ulem} 
\usepackage{array}
\newcolumntype{C}[1]{>{\PreserveBackslash\centering}p{#1}}
\newcolumntype{R}[1]{>{\PreserveBackslash\raggedleft}p{#1}}
\newcolumntype{L}[1]{>{\PreserveBackslash\raggedright}p{#1}}
\usepackage{url}

\ifCLASSOPTIONcompsoc
  \usepackage[nocompress]{cite}
\else
  \usepackage{cite}
\fi
\ifCLASSINFOpdf
\else
\fi
\begin{document}
%
\title{Semantic-Preserving Adversarial Text Attacks}
%
%
%
%

\author{Xinghao~Yang,
        Weifeng~Liu~\IEEEmembership{\textit{Senior Member,~IEEE}},
        James~Bailey,
    Dacheng~Tao~\IEEEmembership{\textit{Fellow,~IEEE}},
        and Wei~Liu~\IEEEmembership{\textit{Senior Member,~IEEE}}
\IEEEcompsocitemizethanks{\IEEEcompsocthanksitem X. Yang and Wei Liu are with the School of Computer Science, University of Technology Sydney, 15 Broadway, Ultimo 2007, NSW, Australia. E-mail: Xinghao.Yang@student.uts.edu.au, Wei.Liu@uts.edu.au.
\IEEEcompsocthanksitem Weifeng Liu is with the School of Information and Control Engineering, China University of Petroleum (East China), Qingdao 266580, China. E-mail: liuwf@upc.edu.cn.
\IEEEcompsocthanksitem J. Bailey is with the Department of Computing and Information Systems, University of Melbourne, Australia. E-mail: baileyj@unimelb.edu.au.
\IEEEcompsocthanksitem D. Tao is with the JD Explore Academy, China, and also with the Digital Science Institute, Faculty of Engineering, The University of Sydney, Darlington,  NSW 2008, Australia  (e-mail: dacheng.tao@gmail.com).}
}

%
%

\markboth
{IEEE TRANSACTIONS ON KNOWLEDGE AND DATA ENGINEERING,~Vol. X, No. X, X~X}
{Shell \MakeLowercase{\textit{}}}
%



\IEEEtitleabstractindextext{%
\begin{abstract}
Deep neural networks (DNNs) are known to be vulnerable to adversarial images, while their robustness in text classification is rarely studied. Several lines of text attack methods have been proposed in the literature, including character-level, word-level, and sentence-level attacks. However, it is still a challenge to minimize the number of word changes necessary to induce misclassification, while simultaneously ensuring  lexical correctness, syntactic soundness, and semantic similarity. In this paper, we propose a Bigram and Unigram based adaptive Semantic Preservation Optimization (BU-SPO) method to examine the vulnerability of deep models. Our method has four major merits. Firstly, we propose to attack text documents not only at the unigram word level but also at the bigram level which better keeps semantics and avoids producing meaningless outputs. Secondly, we propose a hybrid method to replace the input words with options among both their synonyms candidates and sememe candidates, which greatly enriches the potential substitutions compared to only using synonyms. Thirdly, we design an optimization algorithm, i.e., Semantic Preservation Optimization (SPO), to determine the priority of word replacements, aiming to reduce the modification cost. Finally, we further improve the SPO with a semantic Filter (named SPOF) to find the adversarial example with the highest semantic similarity. We evaluate the effectiveness of our BU-SPO and BU-SPOF on IMDB, AG's News, and Yahoo! Answers text datasets by attacking four popular DNNs models. Results show that our methods achieve the highest attack success rates and semantics rates by changing the smallest number of words compared with existing methods.
\end{abstract}

\begin{IEEEkeywords}
Adversarial Machine Learning, Text Attack, Natural Language Processing.
\end{IEEEkeywords}}

\maketitle

\IEEEdisplaynontitleabstractindextext

%
\IEEEpeerreviewmaketitle

\ifCLASSOPTIONcompsoc
\IEEEraisesectionheading{\section{Introduction}\label{sec:introduction}}
\else
\section{Introduction}
\label{sec:introduction}
\fi

%
%
%
%
\IEEEPARstart{D}{eep} neural networks (DNNs) have exhibited brittleness towards adversarial examples primarily in the image domain \cite{szegedy2013intriguing,goodfellow2014explaining}. Adversarial image example can be crafted by intentionally adding a small number of pixel perturbations on the legitimate input. These perturbations are usually hard to be perceived by human vision  but can mislead well-trained DNNs models to erroneous predictions. This phenomenon raises great interest in the image recognition community, and abundant of adversarial attack and defense methods have been proposed to improve the robustnees and interpretability of DNNs \cite{ren2020adversarial}. However, the vulnerability of DNNs in Natural Language Processing (NLP) field is generally underestimated, especially for those security-sensitive NLP tasks, such as spam filtering \cite{9068471}, webpage phishing \cite{he2011efficient}, and sentiment analysis \cite{atallah2001natural}. 

\begin{table}
\small
\centering
\caption{Comparisons between Unigram Attacks and Bigram Attacks. One superiority of Bigram substitution is that it can distinguish commonly used Bigram phrases and avoid generating meaningless sentences.}
\begin{tabular}{l|l|l}
\toprule
\textbf{Original Input} & \textbf{Unigram Attack}  & \textbf{Bigram Attack}  \\
\midrule
New York                & Fresh York     & Empire State \\
Machine Learning                & Device Learning    & Data Mining \\
Primary School          & Major School & Elementary School \\
\bottomrule
\end{tabular}
\label{Tab_bigramExample}
\end{table}

Compared to image attacks, there are non-trivial difficulties in crafting text adversarial samples. Firstly, the text adversarial samples should be lexically correct, syntactically sound, and semantically similar to the original text. This will ensure the adversarial modifications are imperceptible to human readers. Secondly, the words in text sequences are \textit{discrete} tokens instead of continuous pixel values as in images. Therefore, it is infeasible to directly compute the model gradient with respect to every word. Thirdly, making small perturbations on many pixels may still yield a meaningful image from human perception perspectives. However, any small changes, even a single word, to text document can make a sentence meaningless.

\begin{figure*}[t]
\centerline{\includegraphics[width=18.2cm]{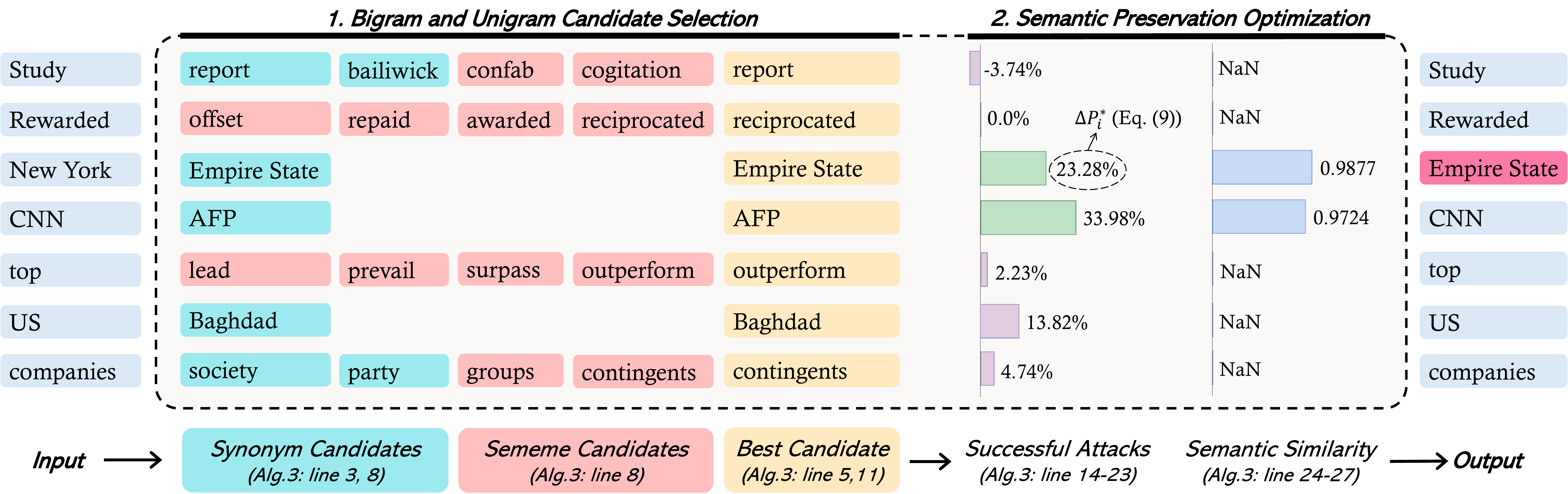}}
\caption{The workflow of our BU-SPOF method with an text example ``Study: CEOs rewarded for outsourcing. NEW YORK (CNN/Money) - The CEOs of the top 50 US companies that sent service jobs overseas pulled down far more pay than their counterparts at other large companies last year, a study said Tuesday." For brevity, in the figure we only show several words from the long text. This example is originally labeled as ``Business" (66.68\%) by LSTM, but is misclassified as ``Sci/Tech" by replacing a bigram ``New York" with ``Empire State". The two green color boxes denote two successful attacks, but we accept the ``Empire State" substitution, because it preserves more semantics (0.9877) than ``AFP" (0.9724).}
\label{fig-flowchart}
\end{figure*}

Several lines of text attack methods have been proposed, such as character-level attack, sentence-level attack, and word-level attack \cite{wang2019towards}. However, character-level attack (e.g., noise $\rightarrow$ nosie) leads to lexical errors, and sentence-level attack (i.e., inserting a whole sentence into the original text) often causes significant semantic changes. To avoid these problems, many recent works focused on word-level attacks that replace the original word with another carefully selected one \cite{zhang2019generating}. However, existing methods mostly generate substitution candidates for every {individual} word (i.e., a unigram), which can easily break commonly used phrases, leading to meaningless outputs (e.g., high school $\rightarrow$ tall school). In addition, when sorting word replacement orders, most algorithms calculate the word importance score (WIS) and attack them via a descending order of the WIS. There are different definitions of WIS, such as probability weighted word saliency (PWWS) \cite{ren2019generating} and the changes of DNNs' predictions before and after deleting a word \cite{jin2019bert}, etc.  One major drawback of using such a static attack order is word substitution inflexibility, e.g., sequentially selecting the top-3 WIS words $\{top1,top2,top3\}$ may not fool a classifier but sometimes the combination $\{top1,top3\}$ can make it.

In this work, we propose a new word-level attack method named Bigram and Unigram based Semantic Preservation Optimization (BU-SPO) which effectively addresses all the drawbacks above. Unlike traditional unigram word attack, we consider both unigram and bigram substitutions. In our approach, we generate more natural candidates by replacing a bigram with its synonyms (e.g., high school $\rightarrow$ secondary school). Table~\ref{Tab_bigramExample} lists several examples that illustrate the superiority of bigram attacks in comparison with unigram attacks. Additionally, we propose to replace input words by considering both their synonym candidates and sememe candidates (i.e., sememe-consistent words). By incorporating these complementary candidates, we have better choices to craft high-quality adversarial texts.

More importantly, we propose an effective candidature search method, Semantic Preservation Optimization (SPO), to determine word replacement priorities. The SPO inherits the best-performed candidate combinations from the previous generation and determines every next replacement word with a heuristic search. For instance, if changing the $\{top1\}$ word cannot mislead a classifier, the static methods used in the literature will select the combination $\{top1, top2\}$ in the second iteration, but our adaptive SPO will check more combinations, e.g., $\{top1, top2\}$, $\{top1, top3\}$, etc. Compared with static strategy, the SPO allows us to fool DNNs models with much fewer modifications, which is significant in reducing grammatical mistakes. In addition, we built a semantic Filter into the SPO algorithm (SPOF), so that it can select the best candidate to maximally preserve the semantic consistency between the input text and the adversarial output. Fig.~\ref{fig-flowchart} illustrates the framework of our algorithm with an attack example. Our main contributions in this work are summarized as below:
\begin{itemize}
\item [1.] We propose to attack text documents not only at the unigram word level but also at the bigram level. This strategy is significant in generating more semantically natural adversarial samples and avoiding meaningless outputs.
\item [2.] We propose a hybrid approach to generate word substitutions from both synonym candidates and sememe candidates. Such a complementary combination provides more options to craft meaningful adversarial examples.
\item [3.] We design an Semantic Preservation Optimization (SPO) method to adaptively determine the word replacement order. The SPO is designed to mislead a DNN classifier using the minimal word modifications compared with static word replacement baselines. Making less word replacements is helpful to reduces the syntactic mistakes. 
\item [4.] We further customize the SPO with a semantic Filter (SPOF), targeting to return the adversarial example that preserves the highest semantic consistency. This step can significantly improve the quality of the output adversarial example in terms of the sentence naturality and fluency.
\item [5.] We conduct extensive experiments on IMDB, AG's News and Yahoo! Answers dataset by attacking both CNN and LSTM models. The experimental results validate the effectiveness of our method in achieving high attack success rate with low perturbation cost and simultaneously keeping high semantic consistency. Besides, our BU-SPOF also shows superiorities in transfer attack, adversarial retrain, and targeted attack compared with baselines.
\end{itemize}

The rest of this paper is organized as follows. In Section \ref{RelatedWorks}, we briefly review the related works in generating text adversarial samples. In Section \ref{algorithm}, we discuss and formalize our algorithm in detail. Evaluation metrics and experimental results are reported in Section \ref{experiments}. Finally, Section \ref{conclusion} concludes this paper.

\section{Related Work}
\label{RelatedWorks}
In this section, we review typical text attack methods and divide them into three groups, including character-level attack \cite{ebrahimi2017hotflip,gao2018black,belinkov2017synthetic}, sentence-level attack \cite{jia2017adversarial,wallace2019universal} and word-level attack \cite{papernot2016crafting,kuleshov2018adversarial,alzantot2018generating,ren2019generating,jin2019bert}.

Firstly, character-level attack \cite{belinkov2017synthetic,ebrahimi2017hotflip,gao2018black} generates adversarial text by deleting, inserting or swapping characters. Belinkov and Bisk \cite{belinkov2017synthetic} devised four types of synthetic
noise: swap, middle random, fully random, and keyboard typo that can mislead the neural machine translation (NMT) models in a large degree. However, they modify every word of an input sentence as they can, which leads to a high perturbation loss. For example, the ``swap" of two letters (e.g., $noise \rightarrow nosie$) is applied to all words with length $\geq 4$, as it does not alter the first and last letters. To reduce the distortion degree, Ebrahimi \textit{et al.} \cite{ebrahimi2017hotflip} proposed HotFlip, which represents every character as a one-hot vector. Then it estimates the best character change by computing directional derivatives with respect to vector operations. Gao \textit{et al.} \cite{gao2018black} designed a black-box DeepWordBug, which evaluates the word importance score by directly removing words one by one and comparing the prediction changes. However, character-level attack breaks the lexical constraint and leads to misspelled word, which can be easily detected and removed by a spell check machine installed before the classifier.

Additionally, sentence-level attack \cite{jia2017adversarial,wallace2019universal} concatenates an adversarial sentence before or more commonly after the clean input text to confuse deep architecture models. For example, Jia and Liang \cite{jia2017adversarial} appended a compatible sentence to the end of paragraph to fool reading comprehension models (RCM). The adversarial sentence looks similar to the original question by combining altered question and fake answers, aiming to mislead RCM into wrong answer location. Nevertheless, this strategy requires a lot of human intervention and cannot be fully automated, e.g., it relies on about 50 manually-defined rules to ensure the adversarial sentence in a declarative form. Recently, Wallace \textit{et al.} \cite{wallace2019universal} sought for the universal adversarial triggers, i.e., input-agnostic sequences, which causes a specific target prediction when it is concatenated to any input from the same data set. The universal sequence is randomly initialized and iteratively updated to increase the likelihood of the target prediction using token replacement gradient as HotFlip. While this method usually leads to dramatic semantic changes and generate human incomprehensible sentences

Finally, word-level attack replaces original input words with carefully picked words. The core problems are (1) how to select proper candidate words and (2) how to determine the word substitution order. Incipiently, Papernot \emph{et al.} \cite{papernot2016crafting} projected words into a 128-dimension embedding space and leveraged the Jacobian matrix to evaluates input-output interaction. However, a small perturbation in the embedding space may lead to totally irrelevant words since there is no hard guarantee that words close in the embedding space are semantically similar. Therefore, subsequent studies focused on synonym substitution strategy that search synonyms from the GloVe embedding space, existing thesaurus (e.g., WordNet and HowNet), or BERT Masked Language Model (MLM). 

By using GloVe, Alzantot \emph{et al.} \cite{alzantot2018generating} designed a population-based genetic algorithm (GA) to imitate the natural selection. However, the GloVe embedding usually fails to distinguish antonyms from synonyms. For example, the nearest neighbors for $expensive$ in GloVe space are $\{pricey, cheaper, costly\}$, where $cheaper$ is its antonyms. Therefore, Glove-based algorithms have to use a counter-fitting method to post-process adversary's vectors to ensure the semantic constraint \cite{mrkvsic2016counter}. Compared with GloVe, utilizing well-organized linguistic thesaurus, e.g., synonym-based WordNet \cite{miller1998wordnet} and sememe-based HowNet \cite{dong2006hownet}, is a simple and easy implementation. Ren \emph{et al.} \cite{ren2019generating} sought synonyms using the WordNet synsets and ranked word replacement order via probability weighted word saliency (PWWS). However, the PWWS sorts word importance score at once and replaces them one by one according to their score descending order. This generally leads to local optimal and word oversubstitution, as the top-k words are not always the strongest combination in order to mislead DNNs models. Zang \emph{et al.} \cite{zang2020word} manifested that the sememe-based HowNet can provide more substitute words than WordNet and proposed the Particle Swarm Optimization (PSO) to determine which group of words should be attacked. In addition, some recent studies utilized BERT MLM to generate contextual perturbations, such as BERT-Attack \cite{li2020bert} and BERT-based Adversarial Examples (BAE) \cite{garg2020bae}. The pre-trained BERT MLM can ensure the predicted token fit in the sentence well, but unable to preserve the semantic similarity. For example, in the sentence ``the food was [MASK]", predicting the [MASK] as $good$ or $bad$ are equally fluent but resulting in opposite sentiment label. Notably, all these works focused on unigram attacks.

\section{Algorithm}
\label{algorithm}
This section details our proposed BU-SPO and BU-SPOF methods. Formally, let $\mathcal{X}=\{\mathbf{X}_1, \mathbf{X}_2, \cdots, \mathbf{X}_N\}$ denote the input space containing $N$ sentences, and $\mathcal{Y}=\{\mathbf{Y}_1, \mathbf{Y}_2, \cdots, \mathbf{Y}_K\}$ represent the output space of $K$ labels. The DNNs classifier $F$ learns a mapping from the text space to the label space $F: \mathcal{X} \rightarrow \mathcal{Y}$.

\subsection{Black-box Text Attack}
\label{DefinitionTextAttack}
We design our method in black-box settings where no network architectures, intermediate parameters or gradient information are available. The only capability of the black-box adversary is to query the output labels (confidence scores) of the threat model, acting as a standard user.

Given a well-trained DNNs classifier $F$, it aims to produce the correct label $\mathbf{Y}_{true} \in \mathcal{Y}$ for any input $\mathbf{X} \in \mathcal{X}$, i.e., $F\left( \mathbf{X} \right) = \mathbf{Y}_{true}$, by maximizing the posterior probability:
\begin{eqnarray}
	\mathop{\arg\max}_{\mathbf{Y}_i \in \mathcal{Y}} P(\mathbf{Y}_i|\mathbf{X}) = \mathbf{Y}_{true}
\end{eqnarray}
A rational text attack  pursues a human-imperceptible perturbation $\Delta\mathbf{X}$ that can fool the classifier $F$ when it is added to the original  $\mathbf{X}$. The altered input $\mathbf{X}^*=\mathbf{X}+\Delta\mathbf{X}$ is defined as the text adversarial example. Generally, a successful adversarial example can  mislead a well-trained classifier into either an arbitrary label other than the true label
\begin{eqnarray}
\label{equ-untargetedAttack}
	\mathop{\arg\max}_{\mathbf{Y}_i \in \mathcal{Y}} P(\mathbf{Y}_i|\mathbf{X}^*) \neq \mathbf{Y}_{true} 
\end{eqnarray}
or a pre-specified label $\mathbf{Y}_{target}$ 
\begin{eqnarray}
\label{equ-targetedAttack}
	\mathop{\arg\max}_{\mathbf{Y}_i \in \mathcal{Y}} P(\mathbf{Y}_i|\mathbf{X}^*) = \mathbf{Y}_{target}
\end{eqnarray}
where $\mathbf{Y}_{target} \neq \mathbf{Y}_{true}$. The attack strategy defined in Eq. (\ref{equ-untargetedAttack}) and Eq. (\ref{equ-targetedAttack}) are known as untargeted attack and targeted attack, respectively. A valid text perturbation needs to satisfy lexical, grammatical, and semantic constraints. As our attack method makes no character modifications, the lexical constraint is naturally retained. Additionally, we propose a bigram substitution strategy to avoid meaningless outputs, and introduce an adaptive search algorithm SPO to minimize the number of word perturbations while preserving the semantic similarity and syntactic coherence.

\subsection{Semantic Similarity}
The semantic similarity between the original input sentence and the adversarial output sentence is vitally important to ensure that the modifications are imperceptible to human. In this paper, we employ the Universal Sentence Encoder (USE) to measure the semantic similarity between text examples \cite{cer2018universal}. The USE model encodes different input sentences into 512 dimensional embedding vectors so that we can easily calculate their cosine similarity score. Specifically, the USE encoder is trained on variety of web text with general purpose, such as Wikipedia, web news, web question-answer pages and discussion forums. Therefore, it is capable of feeding multiple downstream tasks. Formally, denote an USE encoder by $\mathrm{Encoder}$, then the USE score between an example $\mathbf{X}$ and its adversarial variation $\mathbf{X}_{adv}$ is defined as
\begin{eqnarray}
\label{equ-USE-score}
	USE_{score} = Cosine\left(\mathrm{Encoder}(\mathbf{X}), \mathrm{Encoder}(\mathbf{X}_{adv})\right)
\end{eqnarray}
One major advantage of USE sentence embedding is that it can imply how much the selected candidate word fits the original sentence. On the contrary, alternative word embedding methods (e.g., word2vec \cite{rong2014word2vec}), which maps each word to the embedding space, fails to generate context aware representations. 

\subsection{Bigram and Unigram Candidate Selection}
\label{Synonym Candidate Selection}
Before elaborating on the candidate selection procedure, we first briefly introduce the WordNet and HowNet and give the definition of the synonym space and sememe space. WordNet \cite{miller1998wordnet} groups word relations into 117,000 unordered synonym sets (synsets). 
Different synsets are interlinked by super-subordinate relation, e.g., the ``furniture" synset includes the ``bed" synset. In this work, we collect synonym candidates from the WordNet synonym space $\mathbb{W}$.
HowNet \cite{dong2006hownet} annotates words by their sememes, where the sememe is a minimum unit of semantic meaning in linguistics. For example, the word “apple” has multiple sememes, e.g., “fruit”, “computer”, etc. Words sharing the same sememe tag can be interchangeable in crafting adversarial examples. We define the  sememe candidates provided by HowNet as sememe space $\mathbb{H}$.

\subsubsection{Candidate set creation.} Suppose the input sentence contains $n$ words, i.e., $\mathbf{X}=\{ w_1, w_2, \cdots, w_n \}$. For each word $w_i$, we first connect it to its next word $w_{i+1}$ and check if the bigram $(w_i, w_{i+1})$ has synonyms in synonym space $\mathbb{W}$. If yes, we collect all the synonyms to create the bigram candidates set $\mathbb{B}_i$ and skip searching candidates for $w_i$ and $w_{i+1}$ separately\footnote{This means bigram substitution takes precedence.}. Otherwise, we gather all the candidate words for $w_i$ from the synonym space $\mathbb{W}$ and the sememe space $\mathbb{H}$ and denote them as a subset $\mathbb{S}_i \subset \mathbb{W} \cup \mathbb{H}$. It is worth mentioning that we pose a candidate filter here to make sure all the candidate words in $\mathbb{S}_i$ have the same part-of-speech (POS) tags with $w_i$. Replacing words with the same POS tags (e.g., nouns) can help avoid imposing grammatical errors.

If $w_i$ is a named entity (NE), we enlarge the $\mathbb{S}_i$ by absorbing more same-type NE words. The NE refers to a pre-defined real-world object that can be symbolized by a proper noun, such as person names, organizations, and locations \cite{nouvel2016named}. The candidate NE (denoted as $NE_{\mathrm{COMP}}$) must have the same NE type with the original word. It is selected as the most frequently appeared word from the complementary NE set $\mathbb{W}-\mathbb{W}_{\mathbf{Y}_{true}}$ where $\mathbb{W}_{\mathbf{Y}_{true}}$ contains all the NEs of the $\mathbf{Y}_{true}$ class. Then we update the synonym set as $\mathbb{S}_i \leftarrow \mathbb{S}_i \cup NE_{\mathrm{COMP}}$.

Considering polysemy, a word may have more than one sememes defined in HowNet. To guarantee valid substitutions, we take only words that have at least one common sememes with the original word $w_i$ into its candidate set $\mathbb{S}_i$.

\subsubsection{Best candidate selection.} Given the candidate set $\mathbb{S}_i$ (or $\mathbb{B}_i$)\footnote{In the rest of the paper, we slightly abuse the notation by using $\mathbb{S}_i$ to denote the substitution candidate for $w_i$. If $w_i$ belongs to a bigram $(w_i, w_{i+1})$, then $\mathbb{S}_i$ is equivalent to $\mathbb{B}_i$.}, every $w_i' \in \mathbb{S}_i$ is a potential candidate for the replacement of word $w_i$. We define the candidate importance score $I_{w_i'}$ for each substitution candidate $w_i'$ as the reduction of prediction probability:
\begin{eqnarray}
\label{equ-ImportanceScore}
	I_{w_i'} = P(\mathbf{Y}_{true}|\mathbf{X}) - P(\mathbf{Y}_{true}|\mathbf{X}_i'), \forall w_i' \in \mathbb{S}_i
\end{eqnarray}
where
\begin{eqnarray}
	\mathbf{X} = w_1, w_2, \cdots, w_i, \cdots, w_n \\
    \mathbf{X}_i' = w_1, w_2, \cdots, w_i', \cdots, w_n
\end{eqnarray}
Then we pick the word $w_i'$ that achieves the highest $I_{w_i'}$ to be the best substitution word $w_i^*$. Formally, the synonym candidate selection function is as below
\begin{eqnarray}
\label{equ-SynonymSelectionFunction}
	w_i^* = R(w_i, \mathbb{S}_i) = \mathop{\arg\max}_{w_i' \in \mathbb{S}_i} I_{w_i'}
\end{eqnarray}
Repeating this procedure on every word one by one solves the first key issue of our method, as is summarized in  Algorithm~\ref{alg-BU-SPO} from line 1 to line 11.

\begin{algorithm}[t]
\small
\caption{The proposed BU-SPO Algorithm}
\label{alg-BU-SPO}
\KwIn{Sample sentence containing $n$ words $\mathbf{X} = (w_1, \cdots, w_n)$}
\KwIn{Maximum word replacement bond $M$}
\KwIn{Classifier $F$}
\KwOut{Adversarial example $\mathbf{X}_{adv}$}
\tcc{Select candidates for input words}
\For{$i=1$ \KwTo $n$}{
    Connect $w_i$ with its next word as $(w_i,w_{i+1})$; \\
    Collect bigram candidate set $\mathbb{B}_i$ for $(w_i,w_{i+1})$ from synonym space $\mathbb{W}$; \\
    \uIf{$\mathbb{B}_i \neq \varnothing$}{
       Find the best bigram candidate from $\mathbb{B}_i$; \\
       $i \ \ += \ 1$  \Comment*[r]{skip attacking $w_{i+1}$}
       }
    \Else{
         Get a synonym-sememe candidate set $\mathbb{S}_i$ for $w_i$ from $\mathbb{W} \cup \mathbb{H}$; \\
         \If{$w_i$ is a NE}{
          $\mathbb{S}_i \leftarrow \mathbb{S}_i \cup {NE_{\mathrm{COMP}}}$; \\
          } 
          Find the best unigram candidate from $\mathbb{S}_i$; \\
    } 
} 
Create the initial generation with empty $\mathbb{G}^0=\varnothing$;\\
Set the upper bound $M = \min(M, n)$;\\
\tcc{The SPO search starts}
\For{$m=1$ \KwTo $M$}{
    $\mathbb{G}^m = \mathcal{F} \left(\mathbb{G}^{m-1}, \{w_1^*, \cdots,w_n^*\} \right)$ \Comment*[r]{Alg. 2}
    \For{$Candidate \subset \mathbb{G}^m$}{
        Replace $Candidate$ words in $\mathbf{X}$ to craft $\mathbf{X}_{adv}$; \\
        \uIf{$F(\mathbf{X}) \neq F(\mathbf{X}_{adv})$}{
            break \Comment*[r]{successful attack}
            }
        \Else{$\Delta P_{adv} = P(\mathbf{Y}_{true}|\mathbf{X}) - P(\mathbf{Y}_{true}|\mathbf{X}_{adv})$;\\
             } 
        } 
    } 
\Return $\mathbf{X}_{adv}$
\end{algorithm}

\begin{algorithm}[t]
\small
\caption{Function of Generation Creation ($\mathcal{F}$)}
\label{alg-current}
\KwIn{Last generation combinations $\mathbb{G}^{m-1}$}
\KwIn{The best substitution words $\{w_1^*, \cdots, w_n^*\}$}
\KwOut{Current generation combinations $\mathbb{G}^m$}
Initialize current generation $\mathbb{G}^m = \varnothing$; \\
\uIf{$\mathbb{G}^{m-1} = \varnothing$}{
    $\mathbb{G}^m = \{w_1^*, \cdots, w_n^*\}$ \Comment*[r]{$\text{1}^{\text{st}}$ generation}
    }
\Else{
     Search for the most effective element $\mathbb{G}_{best}^{m-1}$ from $\mathbb{G}^{m-1}$ that achieves the highest $\Delta P_{adv}$; \\
     Remove all words of $\mathbb{G}_{best}^{m-1}$ from $\{w_1^* \cdots w_n^*\}$, resulting in $s~(1\leq{s}\leq{n})$ elements left; \\
     \For{$i=1$ \KwTo $s$}{
          Create new generation element $\mathbb{G}^m(i)$ by combing $\mathbb{G}_{best}^{m-1}$ with $i^{th}$ remaining element $w_i^*$; \\
         }
     }
\Return $\mathbb{G}^m$
\end{algorithm}

\subsection{Semantic Preservation Optimization}
\label{Priority Determination}
The Semantic Preservation Optimization (SPO) is designed to determine the word replacement priority with three objectives: 1) achieve a successful attack, 2) make minimal substitutions, and 3) keep the sentence semantic unchanged. Specifically, given the best substitution word $w_i^*$ for the original $w_i$, we obtain $n$ adversarial examples $\{\mathbf{X}_1^*, \cdots, \mathbf{X}_n^*\}$ with each being modified on one word, i.e., $\mathbf{X}_i^* = \{w_1, \cdots, w_i^*, \cdots, w_n\}$. The change of true label probability  between $\mathbf{X}$ and $\mathbf{X}_i^*$ denotes the largest attack effect that can be achieved by modifying $w_i$:
\begin{eqnarray}
\label{equ-bestSynonym}
\Delta P_i^* = P(\mathbf{Y}_{true}|\mathbf{X}) - P(\mathbf{Y}_{true}|\mathbf{X}_i^*)
\end{eqnarray}
A straightforward way of determining the word replacement priority is to sort the words by their $\Delta P_i^*$ in a descent order and select the top-k ones. However, we empirically find that replacing words in such a static order incrementally always leads to local optima and word over substitution. This means simply selecting top-k words using $\Delta P_i^*$ does not necessarily provide the best word combination in misleading DNNs.

In this paper, we propose the Semantic Preservation Optimization (SPO) method, which adaptively determines the word substitution priority. Particularly, we first create the initial generation $\mathbb{G}^0$ as an empty set (line 12 of Algorithm~\ref{alg-BU-SPO}). Then we set the maximum number of words that can be modified, i.e., $M = \min(M, n)$ where $M$ is a predefined replacement cap number. This threshold forces us to stop the loop if the input example does not admit an adversarial alteration after $M$ times of substitution. The SPO procedure is listed between lines 14-21 of Algorithm~\ref{alg-BU-SPO}. For each generation, we first create the population set for the current generation $\mathbb{G}^m$ using the $\mathcal{F}$ function defined in Algorithm~\ref{alg-current}. Specifically, the $\mathcal{F}$ directly returns all the best substitution synonyms $\{w_1^*, \cdots, w_n^*\}$ as the first generation. Then we iteratively query the classifier $F$ and check whether its prediction is changed by replacing the first generation candidates. If a population member $\mathbf{X}_{adv}$ achieves a successful attack, the optimization completes and returns the $\mathbf{X}_{adv}$. Otherwise, we calculate the probability shift of $\Delta P_{adv}$ in line 21. If we can not find a successful attack from the current generation, we proceed to the next iteration while updating the $\Delta P_{adv}$.

In the next generation, we recall the $\mathcal{F}$ to construct $\mathbb{G}^m$ with three steps, as listed in lines 5-8 of Algorithm \ref{alg-current}. Firstly, we search the most effective element from the previous generation $\mathbb{G}^{m-1}$ that attains the maximal $\Delta P_{adv}$. We denote this best element as $\mathbb{G}_{best}^{m-1}$. Then we wipe out all the candidate words belonging to $\mathbb{G}_{best}^{m-1}$ from the full candidates set $\{w_1^*, \cdots, w_n^*\}$, resulting in $s~(1\leq{s}\leq{n})$ elements. Finally, we combine the $\mathbb{G}_{best}^{m-1}$ with every remaining candidate $w_i^*$ and assign it to the current population member $\mathbb{G}^m(i)$. The greedy search between lines 16-21 of Algorithm \ref{alg-BU-SPO} is the same as the first generation but replaces one more word/bigram in every next generation to craft $\mathbf{X}_{adv}$. This procedure does not stop until it successfully finds the adversarial example or reaches the upper threshold of $M$. The SPO method enables us to preserve the best population member from the previous generation and adaptively determine which word should be altered in the current generation. Based on the SPO, we achieve a higher successful attack rate by replacing a fewer number of words compared with static baselines. This solves the second issue.

\subsection{SPO with Semantic Filter (SPOF)}
To further enhance the semantic similarity of SPO, we further improve it by using a semantic filter with the proposed SPOF algorithm (Algorithm \ref{alg-BU-SPOF}). The SPOF employs the same strategy to collect bigram and unigram candidates but improves the word priority determination procedure. Specifically, for each generation, we first create an empty set $SucAdv$ to collect all possible adversarial examples that can make successful attacks (Algorithm \ref{alg-BU-SPOF} line 15). If a population member $\mathbf{X}_{adv}$ achieves a successful attack (Algorithm \ref{alg-BU-SPOF} line 19), we calculate the semantic similarity score between $\mathbf{X}$ and $\mathbf{X}_{adv}$ (Algorithm \ref{alg-BU-SPOF} line 20) and append it into the successful adversarial example set $SucAdv$ (Algorithm \ref{alg-BU-SPOF} line 21). We traverse this procedure for every population member in each generation. Then we find the adversarial example that preserves the highest semantic similarity from the successful adversarial example subset $SucAdv$ if it is not empty (Algorithm \ref{alg-BU-SPOF} line 24-26). Line 27 breaks the optimization to ensure that our SPOF rephrases minimal words in achieving successful attack.

\begin{algorithm}[t]
\small
\caption{The proposed BU-SPOF Algorithm}
\label{alg-BU-SPOF}
\KwIn{Sample sentence containing $n$ words $\mathbf{X} = (w_1, \cdots, w_n)$}
\KwIn{Maximum word replacement bond $M$}
\KwIn{Classifier $F$}
\KwOut{Adversarial example $\mathbf{X}_{adv}$}
\tcc{Select candidates for input words}
\For{$i=1$ \KwTo $n$}{
    Connect $w_i$ with its next word as $(w_i,w_{i+1})$; \\
    Collect bigram candidate set $\mathbb{B}_i$ for $(w_i,w_{i+1})$ from synonym space $\mathbb{W}$; \\
    \uIf{$\mathbb{B}_i \neq \varnothing$}{
       Find the best bigram candidate from $\mathbb{B}_i$; \\
       $i \ \ += \ 1$  \Comment*[r]{skip attacking $w_{i+1}$}
       }
    \Else{
         Get a synonym-sememe candidate set $\mathbb{S}_i$ for $w_i$ from $\mathbb{W} \cup \mathbb{H}$; \\
         \If{$w_i$ is a NE}{
          $\mathbb{S}_i \leftarrow \mathbb{S}_i \cup {NE_{\mathrm{COMP}}}$; \\
          } 
          Find the best unigram candidate from $\mathbb{S}_i$; \\
    } 
} 
Create the initial generation with empty $\mathbb{G}^0=\varnothing$;\\
Set the upper bound $M = \min(M, n)$;\\
\tcc{The SPOF search starts}
\For{$m=1$ \KwTo $M$}{
    $SucAdv = \varnothing$; \\
    $\mathbb{G}^m = \mathcal{F} \left(\mathbb{G}^{m-1}, \{w_1^*, \cdots,w_n^*\} \right)$ \Comment*[r]{Alg. 2}
    \For{$Candidate \subset \mathbb{G}^m$}{
        Replace $Candidate$ words in $\mathbf{X}$ to craft $\mathbf{X}_{adv}$; \\
        \uIf{$F(\mathbf{X}) \neq F(\mathbf{X}_{adv})$}{
            Calculate the $USE_{score}$ between $\mathbf{X}$ and $\mathbf{X}_{adv}$ by Eq. (\ref{equ-USE-score});  \\
            $SucAdv = SucAdv \cup{\mathbf{X}_{adv}}$ \Comment*[r]{successful attack}
            }
        \Else{$\Delta P_{adv} = P(\mathbf{Y}_{true}|\mathbf{X}) - P(\mathbf{Y}_{true}|\mathbf{X}_{adv})$;\\
             } 
        } 
    \tcc{The Semantic Filter}
    \If{$SucAdv \neq \varnothing$}{
        Sort adversarial examples in $SucAdv$ by their $USE_{score}$ descending order; \\
        $\mathbf{X}_{adv} = SucAdv\{0\}$; \\
        break \Comment*[r]{stop the next iterations}
        }
    } 
\Return $\mathbf{X}_{adv}$
\end{algorithm}

\subsection{Targeted Attack Strategy}
Targeted attack is the scenario where attackers aim to misdirect the classifier to a pre-specified target label $\mathbf{Y}_{target}$. In this section, we show our BU-SPO and BU-SPOF algorithms can be easily adapted to conduct targeted attacks by making the following three modifications. Firstly, we change the successful attack condition in Algorithm \ref{alg-BU-SPO} line 18 and Algorithm \ref{alg-BU-SPOF} line 19 from $F(\mathbf{X}) \neq F(\mathbf{X}_{adv})$ to $F(\mathbf{X}_{adv}) = \mathbf{Y}_{target}$. This means we only count adversarial examples that can mislead the classifier to the target label as successful attacks. Secondly, we evaluate the attack strength by calculating how much the target label probability increased rather than how much the true label score decreased. Therefore, the Eq.~(\ref{equ-ImportanceScore}) is reformulated to Eq.~(\ref{equ-ImportanceScore-target})
\begin{eqnarray}
\label{equ-ImportanceScore-target}
	I_{w_i'} = P(\mathbf{Y}_{target}|\mathbf{X}_i') - P(\mathbf{Y}_{target}|\mathbf{X}), \forall w_i' \in \mathbb{S}_i
\end{eqnarray}
and Eq. (\ref{equ-bestSynonym}) is transformed to Eq. (\ref{equ-bestSynonym-target})
\begin{eqnarray}
\label{equ-bestSynonym-target}
\Delta P_i^* = P(\mathbf{Y}_{target}|\mathbf{X}_i^*) - P(\mathbf{Y}_{target}|\mathbf{X})
\end{eqnarray}
Additionally, line 21 of Algorithm \ref{alg-BU-SPO} and line 23 of Algorithm~\ref{alg-BU-SPOF} become Eq. (\ref{equ-target})
\begin{eqnarray}
\label{equ-target}
\Delta P_{adv} = P(\mathbf{Y}_{target}|\mathbf{X}_{adv}) - P(\mathbf{Y}_{target}|\mathbf{X}).
\end{eqnarray}
Finally, we select the most frequent NE substitution from the target class, i.e., $NE_{Target}$, instead of from the complementary set $NE_{\mathrm{COMP}}$. This is helpful to increase the target label score and improve the success rate of targeted attacks.

\section{Experiments}
\label{experiments}
We evaluate the effectiveness of our BU-SPO and BU-SPOF methods on widely used text datasets. We provide code and data with a Github repository\footnote{ \url{https://github.com/AdvAttack/BU-SPO}} to ensure reproducibility.

\begin{table}
\centering
\caption{Dataset information summarization. ``\# Avg. Words" is the average number of words for all samples.}
\begin{tabular}{lccc}
\toprule
Dataset       &  IMDB     &  AG's News  &  Yahoo! Answers \\
\midrule
\# Train      &  2,5000   &  120,000    &  1,400,000      \\
\# Test       &  2,5000   &  7,600      &  60,000         \\
\# Avg. Words &  227.11   &  37.84      &  32.39          \\
\# Classes    &  2        &  4          &  10             \\
\bottomrule
\end{tabular}
\label{Tab_dataset}
\end{table}

\begin{table}
\centering
\caption{Test Accuracy of Four DNNs Models before Attacks.}
\begin{tabular}{lm{1.6cm}<{\centering}m{2.8cm}<{\centering}}
\toprule
Dataset                           &  Model         &  Original Test Accuracy  \\
\midrule
~                                 &  CNN           &  87.97\%    \\
IMDB                              &  LSTM          &  87.94\%    \\
~                                 &  Bi-LSTM       &  85.71\%    \\
\midrule
~                                 &  CNN           &  90.75\%    \\
AG's News                         &  LSTM          &  91.62\%    \\
~                                 &  Ch-CNN        &  89.24\%    \\
\midrule
~                                 &  CNN           &  71.21\%    \\
Yahoo! Answers                    &  Bi-LSTM       &  71.60\%    \\
~                                 &  Ch-CNN        &  64.62\%    \\
\bottomrule
\end{tabular}
\label{Tab_accuracy}
\end{table}

\begin{table*}[t]
\small
\centering
\caption{The Attack Success Rate (ASR) of various attack algorithms. For each row, the highest ASR is highlighted in bold, the second highest ASR is highlighted in underline, and the third highest ASR is denoted with italic font.}
\begin{tabular}{p{2.5cm}|m{1.2cm}<{\centering}m{1.3cm}<{\centering}m{1.2cm}<{\centering}|m{1.2cm}<{\centering}m{1.2cm}<{\centering}m{1.3cm}<{\centering}|m{1.2cm}<{\centering}m{1.3cm}<{\centering}m{1.3cm}<{\centering}}
\toprule
\multirow{2}[3]{*}{Methods} &  \multicolumn{3}{c|}{IMDB} & \multicolumn{3}{c|}{AG's News}   &  \multicolumn{3}{c}{Yahoo! Answers}     \\
\cmidrule(lr){2-4} \cmidrule(lr){5-7} \cmidrule(lr){8-10}
~         & CNN             & Bi-LSTM    & LSTM       & CNN        & LSTM       & Ch-CNN    & CNN      & Bi-LSTM    & Ch-CNN     \\
\midrule
BEAT      & $91.02\%$       & $90.15\%$  & \underline{99.77\%}  & $86.91\%$  & $77.52\%$  & N/A         & N/A      & N/A        & N/A          \\
PSO       & \textbf{100\%}  & \textbf{100\%}  & \textbf{100\%} & $85.48\%$  & $79.87\%$   & N/A         & N/A      & N/A        & N/A   \\
TEFO      & \textbf{100\%}  & \textbf{100\%}  & \textbf{100\%} & $82.75\%$  & $85.17\%$   & N/A         & N/A      & N/A        & N/A   \\
PWWS      & \textit{94.60\%}& \underline{99.76\%}  & $90.01\%$  & $85.48\%$  & $79.87\%$  & $75.48\%$   & $68.39\%$  & $67.76\%$  & $97.59\%$  \\
WSA       & $38.85\%$       & $81.70\%$  & $31.65\%$  & $77.98\%$  & $74.73\%$  & $68.33\%$   & $66.82\%$  & $66.20\%$  & $96.36\%$  \\
PWWS ($M=20$)  & $90.56\%$       & $97.09\%$  & $82.60\%$  & $81.98\%$  & $76.81\%$  & $75.25\%$   & $68.23\%$  & $67.29\%$  & $96.54\%$  \\
RAND      & $81.13\%$       & $94.79\%$  & $74.33\%$  & $82.42\%$  & $79.76\%$  & $75.26\%$   & $51.96\%$  & $51.64\%$  & $95.50\%$  \\
\midrule
U-SPO    & \underline{95.58\%} & \textit{98.90\%}  & $90.80\%$  & $85.76\%$  & $81.07\%$  & \textit{80.70\%}   & $70.27\%$  & $67.92\%$  & $97.59\%$ \\
HU-SPO   & \textbf{100\%}  & \textbf{100\%} & \textit{98.65\%}   & \underline{92.77\%}  & \underline{88.40\%}  & \textbf{92.17\%}   & \textbf{90.14\%}  & \textbf{87.95\%}  & \textbf{98.91\%} \\
BU-SPO   & \textbf{100\%}  & \textbf{100\%} & $98.32\%$ & \textit{91.77\%} & \textit{86.21\%}  & \underline{91.49\%}   & \underline{88.58\%}  & \underline{85.60\%}  & \underline{98.50\%}  \\
BU-SPOF    & \textbf{100\%}  & \textbf{100\%}  & \textbf{100\%} & \textbf{93.44\%}  & \textbf{89.06\%}   & \underline{91.49\%} & \textit{87.17\%} & \textit{85.45\%} & \textit{98.29\%}   \\
\bottomrule
\end{tabular}
\label{Tab_MeanFoolingRate}
\end{table*}

\begin{table*}[t]
\small
\centering
\caption{The Average Word Replacement (AWR) number of various attack methods. For each row, the smallest AWR is highlighted in bold, the second smallest AWR is denoted in underline, and the third smallest AWR is represented with italic font.}
\begin{tabular}{p{2.5cm}|m{1.2cm}<{\centering}m{1.3cm}<{\centering}m{1.2cm}<{\centering}|m{1.2cm}<{\centering}m{1.2cm}<{\centering}m{1.3cm}<{\centering}|m{1.2cm}<{\centering}m{1.3cm}<{\centering}m{1.3cm}<{\centering}}
\toprule
\multirow{2}[3]{*}{Methods} &  \multicolumn{3}{c|}{IMDB} & \multicolumn{3}{c|}{AG's News}   &  \multicolumn{3}{c}{Yahoo! Answers}     \\
\cmidrule(lr){2-4} \cmidrule(lr){5-7} \cmidrule(lr){8-10}
~         & CNN             & Bi-LSTM    & LSTM       & CNN        & LSTM       & Ch-CNN    & CNN      & Bi-LSTM    & Ch-CNN     \\
\midrule
BEAT      & $7.50$          & $7.89$     & $9.88$     & $5.97$     & $6.24$     & N/A         & N/A      & N/A        & N/A     \\
PSO       & $3.42$          & $5.22$     & $5.91$     & $5.02$     & $5.82$     & N/A         & N/A      & N/A        & N/A     \\
TEFO      & $8.01$          & $8.13$     & $7.43$     & $7.42$     & $8.49$     & N/A         & N/A      & N/A        & N/A     \\
PWWS      & $8.63$         & $5.53$     & $14.02$    & $6.29$     & $8.05$     & $5.59$      & $3.34$   & $3.47$     & $1.05$  \\
WSA       & $16.15$         & $10.55$    & $16.85$    & $9.51$     & $10.12$    & $8.23$      & $3.95$   & $3.83$     & $1.21$  \\
PWWS ($M=20$)      & $5.87$          & $5.16$     & $7.84$     & $6.03$     & $7.72$     & $5.44$      & $3.15$   & $3.20$      & $1.23$  \\
RAND      & $7.31$          & $5.67$     & $8.79$     & $5.23$     & $6.19$     & $5.06$      & $3.65$   & $3.74$     & $1.12$  \\
\midrule
U-SPO    & $4.50$           & $4.24$    & $6.21$     & $4.82$    & \textit{5.97}     & $4.39$     & $2.78$   & $2.98$     & \textit{1.04}  \\
HU-SPO   & \underline{2.07}& \underline{2.28} & \textit{2.67}     & \textbf{4.16}     & \underline{5.74}     & \textbf{3.41}   & \textbf{1.85}   & \textbf{2.28}     & \textbf{1.02}  \\
BU-SPO   & \textit{2.11}          & \textit{2.33}     & \underline{2.58}     & \textit{4.32}     & $5.99$     & \underline{3.52}      & \underline{1.89}   & \textit{2.40}      & \textit{1.04}  \\
BU-SPOF    & \textbf{2.06}   & \textbf{1.62} & \textbf{2.38}  & \underline{4.17}     & \textbf{5.65}  & \textit{3.62}      & \textit{1.96}   & \underline{2.32}     & \underline{1.03}  \\
\bottomrule
\end{tabular}
\label{Tab_MeanSubstitutionWords}
\end{table*}

\begin{table*}[t]
\small
\centering
\caption{The average Universal Sentence Encoder (USE) score of various attack methods. For each row, the highest USE score is highlighted in bold, the second highest USE score is denoted in underline, and the third highest USE score is represented with italic.}
\begin{tabular}{p{2.5cm}|m{1.2cm}<{\centering}m{1.3cm}<{\centering}m{1.2cm}<{\centering}|m{1.2cm}<{\centering}m{1.2cm}<{\centering}m{1.3cm}<{\centering}|m{1.2cm}<{\centering}m{1.3cm}<{\centering}m{1.3cm}<{\centering}}
\toprule
\multirow{2}[3]{*}{Methods} &  \multicolumn{3}{c|}{IMDB} & \multicolumn{3}{c|}{AG's News}   &  \multicolumn{3}{c}{Yahoo! Answers}     \\
\cmidrule(lr){2-4} \cmidrule(lr){5-7} \cmidrule(lr){8-10}
~         & CNN             & Bi-LSTM    & LSTM       & CNN        & LSTM       & Ch-CNN    & CNN      & Bi-LSTM    & Ch-CNN     \\
\midrule
BEAT      & $0.9825$        & $0.9764$   & $0.9786$   & $0.859$    & $0.8412$   & N/A         & N/A      & N/A        & N/A     \\
PSO       & \textit{0.9940}  & $0.9889$   & $0.9849$   & \textbf{0.9175}   & \textbf{0.9056}   & N/A         & N/A      & N/A        & N/A     \\
TEFO      & $0.9878$        & $0.9851$   & \textit{0.9859}   & $0.8773$   & $0.8530$    & N/A         & N/A      & N/A        & N/A     \\
PWWS      & $0.9905$        & $0.9852$   & $0.9660$    & $0.8557$   & $0.8079$   & $0.9223$    & $0.8012$ & $0.7915$   & $0.9019$ \\
WSA       & $0.9514$        & $0.9325$   & $0.9192$   & $0.7770$    & $0.7565$   & $0.8481$    & $0.7558$ & $0.7542$   & $0.8965$ \\
PWWS ($M=20$)      & $0.9907$        & $0.9868$   & $0.9761$   & $0.8593$   & $0.8206$   & $0.9241$    & $0.8014$ & $0.7923$   & $0.8993$ \\
RAND      & $0.9893$        & $0.9839$   & $0.9711$   & $0.8874$   & $0.8519$   & $0.9326$    & \textit{0.8045} & \textit{0.7940}    & \textit{0.8994} \\
\midrule
U-SPO    & $0.9921$        & $0.9881$   & $0.9804$   & $0.8928$   & $0.8585$   & $0.9350$     & \underline{0.8176} & \underline{0.8048}   & \underline{0.9020}  \\
HU-SPO   & \underline{0.9945}        & \textit{0.9934}       & \underline{0.9922}   & $0.9028$   & $0.8622$   & \textit{0.9416}    & $0.7969$ & $0.7735$   & $0.8861$ \\
BU-SPO   & \underline{0.9945}        & \underline{0.9935}   & \underline{0.9922}   & \textit{0.9044}   & \textit{0.8647}   & \underline{0.9431}    & $0.8017$ & $0.7752$   & $0.8877$ \\
BU-SPOF  & \textbf{0.9970}  & \textbf{0.9958}   & \textbf{0.9941}   & \underline{0.9122}   & \underline{0.8749}   & \textbf{0.9526}    & \textbf{0.8292} & \textbf{0.8108}   & \textbf{0.9344} \\
\bottomrule
\end{tabular}
\label{Tab_MeanUSEScore}
\end{table*}

\subsection{Datasets}
\label{Datasets and victim models}
We conduct experiments on three publicly available benchmarks, including IMDB, AG's News, and Yahoo! Answers. Details of these datasets are summarized in Table~\ref{Tab_dataset}.

IMDB \cite{maas2011learning} is a binary sentiment classification dataset containing 50,000 movie reviews, where 25,000 samples are used for training and 25,000 for testing. The average text length is 227 words (without punctuation). 
AG's News \cite{zhang2015character} is a news classification dataset with 4 topic classes, i.e., World, Sports, Business and Sci/Tech. Each class consists of 30,000 train examples and 1,900 test documents. It totally contains 120,000 training sample and 7,600 testing samples. 
Yahoo! Answers \cite{zhang2015character} is a big topic classification dataset with 1,400,000 text samples for training and 60,000 for testing. These examples belong to 10 topic categories.

\subsection{Victim Models}
\label{sec-victim-models}
We apply our attack algorithm on four popular victim models, including two Convolutional Neural Networks (CNNs) and two Recurrent Neural Networks (RNNs). These popular models are effective tools for text classification in either word level or character level.

Word-based CNN (\textbf{CNN}) \cite{kim-2014-convolutional} is stacked by a word embedding layer with 50 embedding dimensions, a convolutional layer with 250 filters
, a global max pooling layer, two pairs of fully-connected layer and nonliner activation layer. Besides, it contains two dropout layers with 0.2 dropout rate to prevent overfitting. This Word CNN model is implemented on all the three datasets.

Character-based CNN (\textbf{Ch-CNN}) \cite{zhang2015character} is composed of a 69-dimensional character embedding layer, 6 convolutional layers and 3 densely-connected layers. Each convolutional layer employs 256 filter kernels with filter size varying from 3 to 7. It also inserts one dropout layer after every densely-connected layer with 0.1 dropout frequency. We evaluate this Ch-CNN model on AG's News dataset.

Word-based LSTM (\textbf{LSTM}) passes the input sequence through a 100-dimension embedding layer, concatenating a 128-units long short-term memory layer, followed by a 0.5 fraction dropout layer. The LSTM structure prevents gradient vanishing by utilizing a memory cell, and is effective for sequence text classification \cite{papernot2016crafting}. This Word LSTM model is applied on the AG's News dataset.

Bidirectional LSTM (\textbf{Bi-LSTM}) consists of a 128-dimension word embedding layer, a bidirectional layer that wraps 64 LSTM units. Then it combines a 0.5 proportion dropout layer and a fully-connected layer for classification. We run this bidirectional LSTM model on both IMDB and Yahoo! Answer datasets.

Table~\ref{Tab_accuracy} lists the classification accuracy of these models on the original legitimate test samples.

\subsection{Evaluation Metrics}
We use three methods to evaluate the text attack performance, i.e., Attack Success Rate (ASR), the average word replacement (AWR) number, and the semantic similarity score (USE score). 

The ASR indicates how much an adversary can mislead the victim model. Formally, an successful attack is when the classifier $F$ can correctly classify the original legitimate input $F\left( \mathbf{X} \right) = \mathbf{Y}_{true}$ but makes a wrong prediction on the corresponding attacked input $F\left( \mathbf{X}+\Delta\mathbf{X} \right) = \mathbf{Y}^*$. Therefore, the ASR is defined as 
\begin{eqnarray}
\text{ASR} = \frac{\sum_{\mathbf{X}\in\mathcal{X}}\{ F(\mathbf{X})= \mathbf{Y}_{true} \wedge F(\mathbf{X}+\Delta\mathbf{X})= \mathbf{Y}^* \}}{\sum_{\mathbf{X}\in \mathcal{X}} \{F(\mathbf{X})= \mathbf{Y}_{true} \}}
\end{eqnarray}
where $\mathbf{Y}^*$ can be any label different from the $\mathbf{Y}_{true}$ (untargeted attack) or equal to a user specified label (targeted attack). The $\Delta\mathbf{X}$ denotes the modifications for the legitimate text sample. To accurately quantify the perturbation loss, we straightforwardly count the number of word modifications for each sample and compute the average word replacement (AWR) number to denote the adversarial attack cost. The semantic similarity score is also an important indicator to evaluate the quality of the crafted adversarial, as it measures whether the adversarial example reads natural and fluent. Intuitively, a rational hacker hopes to attain high ASR and semantic similarity score while modifying a small number of words.

\subsection{Baselines}
\label{baselines}
We compare our method with representative black-box word-level attack algorithms as listed below.

\begin{itemize}
\item RAND (attack randomly) selects a synonym from WordNet and ranks the attack order by our SPO algorithm.
\item Word saliency attack (WSA) \cite{li2016understanding} gets replacement words from WordNet and rephrases texts in the word saliency (WS) descending order. The word saliency is similar to Eq. (\ref{equ-ImportanceScore}) but replaces $w_i$ with $unknown$.
\item PWWS \cite{ren2019generating} chooses candidate words from WordNet and sorts word attack order by multiplying the word saliency and probability variation.
\item PSO \cite{zang2020word} selects word candidates from HowNet and employs the PSO to find adversarial text. This method treats every sample as a particle where its location in the search space needs to be optimized.
\item TextFooler (TEFO) \cite{jin2019bert} obtains synonyms from Glove space and defines the WIS by iteratively deleting input words and calculating the DNNs score changes.
\item BERT-ATTACK (BEAT) \cite{li2020bert} takes advantage of BERT MLM to generate candidates and attack words by the static WIS descending order. The WIS is similar to Eq. (\ref{equ-ImportanceScore}) but changes $w_i$ to a masked word.
\end{itemize}

\subsection{Experimental Settings} 
We train all the DNNs models using the ADAM optimizer \cite{kingma2014adam}, where parameters are set as: learning rate $=0.001$, $\beta_1=0.9$, $\beta_2=0.999$, $\epsilon=10^{-7}$. We deploy BU-SPO and the first three baselines on Keras. The PSO, TEFO, and BEAT are tested on the TextAttack framework \cite{morris2020textattack}, where the Ch-CNN model and the Yahoo! Answers are currently unavailable. For this reason, results under these settings are shown as infeasible to obtain (i.e., N/A) in Table \ref{Tab_MeanFoolingRate}, Table \ref{Tab_MeanSubstitutionWords}, and Table \ref{Tab_MeanUSEScore}. We set the upper bond of word replacement number as $M=20$ for our methods. This means we will stop the attack iteration if a text sample does not admit the adversarial attack after 20 substitutions. For the baseline methods, we use their recommended parameters for fair comparison. Particularly, for the most related baseline PWWS, we also report its performance with the constraint, i.e., $M=20$, for fair comparison. To achieve efficiency, their attack performance is assessed on 1000 test samples of each dataset as the conventional setting \cite{zang2020word,jin2019bert}. 

\begin{table*}[t]
\small
\centering
\caption{Adversarial examples of IMDB (attack Word CNN). Green texts are original words, while red ones are substitutions.}
\begin{tabular}{p{17.5cm}}
\toprule
\multicolumn{1}{c}{IMDB Example 1} \\
\toprule
PWWS (Successful attack. True label score: $70.5\% \rightarrow 1.21\%$) \\
\midrule
Nikki Finn is the \sout{{\color{blue} kind}} {\color{red} variety} of girl I would marry. Never boring, always thinking positively, good with animals. Okay, as one reviewer wrote, a bit too much peroxide, lipstick, and eyebrows (Only Madonna could get away with that). But that's why I love Nikki Finn, she's not your ordinary girl. She makes things happen, always exciting to be around, and always honest. 
\\
\toprule
BU-SPOF (Successful attack. True label score: $70.5\% \rightarrow 1.91\%$) \\
\midrule
Nikki Finn is the kind of girl I would marry. Never boring, always thinking positively, good with animals. Okay, as one reviewer wrote, \sout{{\color{blue} a bit}} {\color{red} a little} too much peroxide, lipstick, and eyebrows (Only Madonna could get away with that). But that's why I love Nikki Finn, she's not your ordinary girl. She makes things happen, always exciting to be around, and always honest. 
\\
\toprule
\multicolumn{1}{c}{IMDB Example 2} \\
\toprule
PWWS (Successful attack. True label score: $97.41\% \rightarrow 43.81\%$) \\
\midrule
\sout{{\color{blue} Alfred}} {\color{red} Lee} \sout{{\color{blue} Hitchcock}} {\color{red} Lee} shows originality in the remake of his own 1934 British film, "The Man Who Knew Too Much". This 1956 take on the same story is much lighter than the previous one. Mr. Hitchcock was lucky in having collaborators that went with him from one film to the next, thus keeping a standard in his work. Robert Burks did an \sout{{\color{blue} excellent}} {\color{red} splendid} \sout{{\color{blue} job}} {\color{red} problem} with the cinematography and George Tomasini's editing shows his talent. Ultimately, Bernard Herrmann is seen conducting at the magnificent Royal Albert Hall in London at the climax of the picture. James Stewart was an actor that worked well with Mr. Hitchcock. In this version, he plays a doctor from Indiana on vacation with his wife and son. When we meet him, they are on their way to Marrakesh in one local bus and the intrigue begins. 
\\
\toprule
BU-SPOF (Successful attack. True label score: $97.41\% \rightarrow 47.84\%$) \\
\midrule
\sout{{\color{blue} Alfred Hitchcock}} {\color{red} Alfred Joseph Hitchcock} shows originality in the remake of his own 1934 British film, "The Man Who Knew Too Much". This 1956 take on the same story is much lighter than the previous one. Mr. Hitchcock was lucky in having collaborators that went with him from one film to the next, thus keeping a standard in his work. Robert Burks did an excellent \sout{{\color{blue} job}} {\color{red} duty} with the cinematography and George Tomasini's editing shows his talent. Ultimately, Bernard Herrmann is seen conducting at the magnificent Royal Albert Hall in London at the climax of the picture. James Stewart was an actor that worked well with Mr. Hitchcock. In this version, he plays a doctor from Indiana on vacation with his wife and son. When we meet him, they are on their way to Marrakesh in one local bus and the intrigue begins.
\\
\bottomrule
\end{tabular}
\label{Tab_imdbExample}
\end{table*}

\begin{table*}[t]
\small
\centering
\caption{Adversarial examples by attacking Word LSTM model on AG's News dataset.}
\begin{tabular}{p{17.5cm}}
\toprule
\multicolumn{1}{c}{AG's New Example 1} \\
\toprule
PWWS (Successful attack. True label score: $90.86\% \rightarrow 37.17\%$) \\
\midrule
Afghan women \sout{{\color{blue} make}} {\color{red} arrive} brief Olympic \sout{{\color{blue} debut}} {\color{red} introduction}. Afghan women made a short-lived debut in the Olympic Games on Wednesday as 18-year-old judo wildcard Friba Razayee was defeated after 45 seconds of her first \sout{{\color{blue} match}} {\color{red} peer} in the under-70kg middleweight.  \\
\toprule
BU-SPOF (Successful attack. True label score: $90.86\% \rightarrow 29.69\%$) \\
\midrule
Afghan women make brief Olympic debut. Afghan women made a short-lived debut in the \sout{{\color{blue} Olympic Games}} {\color{red} Olympiad} on Wednesday as 18-year-old judo wildcard Friba Razayee was defeated after 45 seconds of her first match in the under-70kg middleweight.  \\
\toprule
\multicolumn{1}{c}{AG's News Example 2} \\
\toprule
PWWS (Successful attack. True label score: $66.13\% \rightarrow 45.21\%$) \\
\midrule
Internosis Will Relocate To Greenbelt in October.  Internosis Inc., an \sout{{\color{blue} information}} {\color{red} entropy} technology company in Arlington, plans to move its headquarters to Greenbelt in October. The relocation will bring 170 jobs to Prince George's County. \\
\toprule
BU-SPOF (Successful attack. True label score: $97.41\% \rightarrow 18.42\%$) \\
\midrule
Internosis Will Relocate To Greenbelt in October.  Internosis Inc., an \sout{{\color{blue} information technology}} {\color{red} IT} company in Arlington, plans to move its headquarters to Greenbelt in October. The relocation will bring 170 jobs to Prince George's County. \\
\bottomrule
\end{tabular}
\label{Tab_agnewsExample}
\end{table*}

\begin{table*}[t]
\small
\centering
\caption{Adversarial examples by attacking Bi-LSTM model on Yahoo! Answers dataset.}
\begin{tabular}{p{17.5cm}}
\toprule
\multicolumn{1}{c}{Yahoo! Answers Example 1} \\
\toprule
PWWS (Failure. True label score: $92.54\% \rightarrow 43.65\%$) \\
\midrule
What \sout{{\color{blue} are}} {\color{red} exist} \sout{{\color{blue} good}} {\color{red} honorable} resources to \sout{{\color{blue} learn}} {\color{red} memorize} about treatments for prostate cancer? \\
\toprule
BU-SPOF (Successful attack. True label score: $92.54\% \rightarrow 30.42\%$) \\
\midrule
What are good resources to learn about treatments for \sout{{\color{blue} prostate cancer}} {\color{red} prostatic adenocarcinoma}? \\
\toprule
\multicolumn{1}{c}{Yahoo! Answers Example 2} \\
\toprule
PWWS (Successful attack. True label score: $81.52\% \rightarrow 40.85\%$) \\
\midrule
Why did president \sout{{\color{blue} Bush}} {\color{red} Equine} get his Masters degree?  \\
\toprule
BU-SPOF (Successful attack. True label score: $81.52\% \rightarrow 2.37\%$) \\
\midrule
Why did \sout{{\color{blue} president Bush}} {\color{red} Dubyuh} get his Masters degree? \\
\bottomrule
\end{tabular}
\label{Tab_yahooExample}
\end{table*}

\subsection{Experimental Results and Analysis}
\label{ExperimentalResults}
The experimental results of ASR, AWR, and the semantic similarity score are listed in Table~\ref{Tab_MeanFoolingRate} and Table~\ref{Tab_MeanSubstitutionWords}, and Table~\ref{Tab_MeanUSEScore}, respectively. We manifest the first four contributions mentioned in the Introduction by asking four research questions:

\noindent\textbf{Q1: Is our adaptive SPO superior to static baselines?} To validate this, we design the U-SPO that searches substitution words from \textit{only} WordNet and attacks text \textit{only} at the unigram word level - the same as WSA and PWWS, but employs our SPO to determine the word substitution priority. Experimental results in Table~\ref{Tab_MeanFoolingRate} and Table~\ref{Tab_MeanSubstitutionWords} show that U-SPO achieves higher ASR and changes a much smaller number of words comparing with static counterparts (WSA, and PWWS). Besides, the RAND delivers higher SAR than WSA on IMDB and AG's News. This also illustrates the merit of our adaptive SPO.

\noindent\textbf{Q2: Is the hybrid of synonym and sememe beneficial?} We present a Hybrid version of U-SPO, i.e., HU-SPO, which is all the same with U-SPO but integrates HowNet to search synonym-sememe candidates. Table~\ref{Tab_MeanFoolingRate} shows that HU-SPO  accomplishes the highest ASR in most cases and outperforms U-SPO by a large margin. Intriguingly, Table~\ref{Tab_MeanSubstitutionWords} exhibits that HU-SPO achieves such high ASR by using fewer word substitutions. This strongly suggests the profit of incorporating HowNet in the candidate selection step.

\noindent\textbf{Q3: What's the advantage of combining the bigram attack?} The bigram substitution is vitally significant in improving semantic smoothness and generating meaningful sentences. To show this, we propose the BU-SPO method, which considers both bigram and unigram attack. Compared with HU-SPO, the BU-SPO achieves higher USE score even if it changes more words. This means bigram substitution can avoid produce meaningless sentences. In addition, we list two adversarial examples from IMDB (Table~\ref{Tab_imdbExample}), AG's News (Table~\ref{Tab_agnewsExample}) and Yahoo! Answers (Table~\ref{Tab_yahooExample}) for qualitative analysis. We can see from the adversarial examples that our bigram substitution can greatly reduce the semantic variations. For example, Table \ref{Tab_agnewsExample}  shows that our method replaces two words (i.e., information technology $\rightarrow$ IT) but causes less semantic variation than PWWS only changing one word (information $\rightarrow$ entropy).

\noindent\textbf{Q4: Can the semantic filter really improve semantic similarity?} A straightforward way to validate this point is to compare the BU-SPO and BU-SPOF, since the only difference between the two algorithms is whether they use semantic filter. From Table \ref{Tab_MeanUSEScore} we can see that BU-SPOF attains higher semantic similarity than BU-SPO and often achieves the highest USE score compared with all baselines. This confirms our expectation that the semantic filter is significant in improving the naturality and fluency of the generated adversarial examples.

Overall, Table~\ref{Tab_MeanFoolingRate}, Table~\ref{Tab_MeanSubstitutionWords}, and Table~\ref{Tab_MeanUSEScore} elaborate that our proposed algorithms (U-SPO, HU-SPO, BU-SPO, and BU-SPOF) almost swept the top-3 results on all datasets and victim models, indicating the superiority of our method.

\subsection{Transferability}
\label{transferability}
\begin{figure}[t]
\centerline{\includegraphics[width=8.5cm]{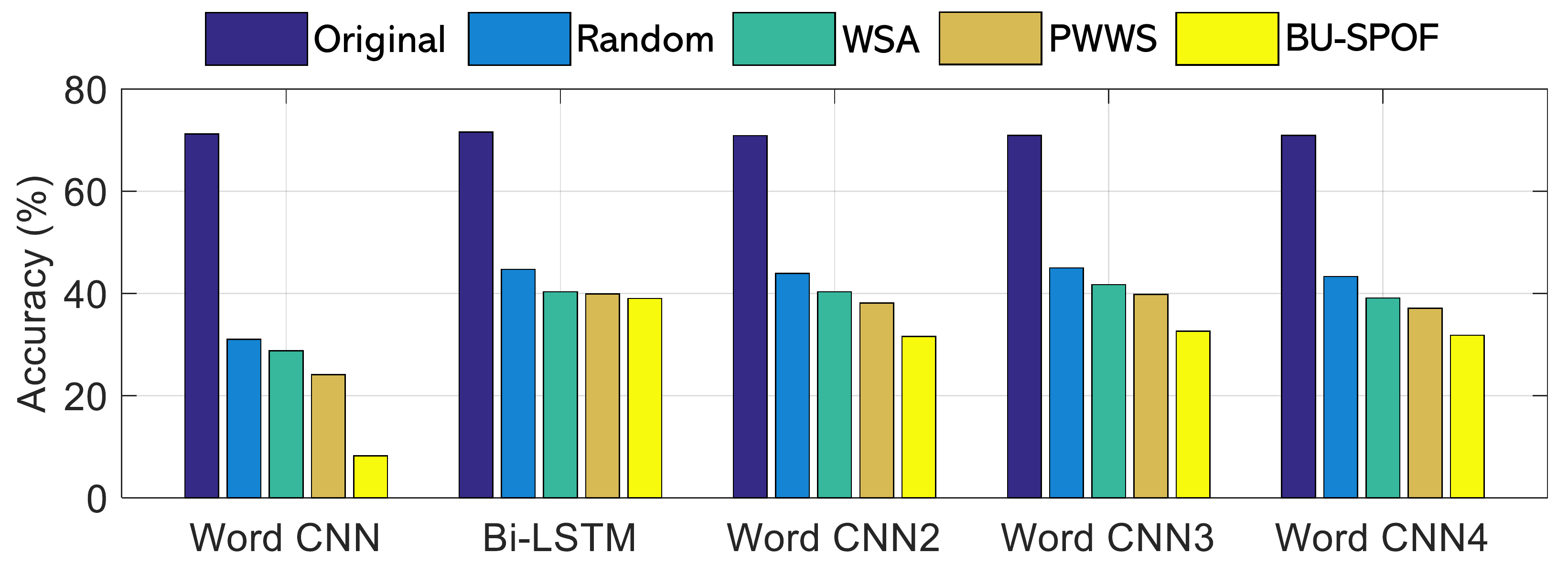}}
\caption{Transfer attack on Yahoo! Answers. Lower accuracy indicates higher transfer ability  (the lower the better).}
\label{fig-transfer}
\end{figure}

Transferability of adversarial examples is the ability that the adversarial samples generated to mislead a specific model $F$ can also be used to mislead other well-trained models $F'$ - even if their network structures greatly differ \cite{papernot2016transferability}. To evaluate whether our adversarial samples are transferable between models, we construct three more CNN models named Word CNN2, Word CNN3 and Word CNN4. Different from the previous Word CNN model (described in section \ref{sec-victim-models}), Word CNN2 has one more fully connected layer, the Word CNN3 replaces the Relu nonlinear function with Tanh, and Word CNN4 adds one convolutional layer. We apply the 1000 adversarial examples generated on Word CNN to attack Word CNN2, Word CNN3, Word CNN4, and the LSTM model. Fig.~\ref{fig-transfer} shows the results on the original Word CNN and transferred models. It can be seen from Fig.~\ref{fig-transfer} that our method attains the best transfer attack performance, elaborating the strength of our method in transfer attack.

\subsection{Adversarial Retraining}
\label{retrain}
\begin{figure}[t]
\centerline{\includegraphics[width=7cm]{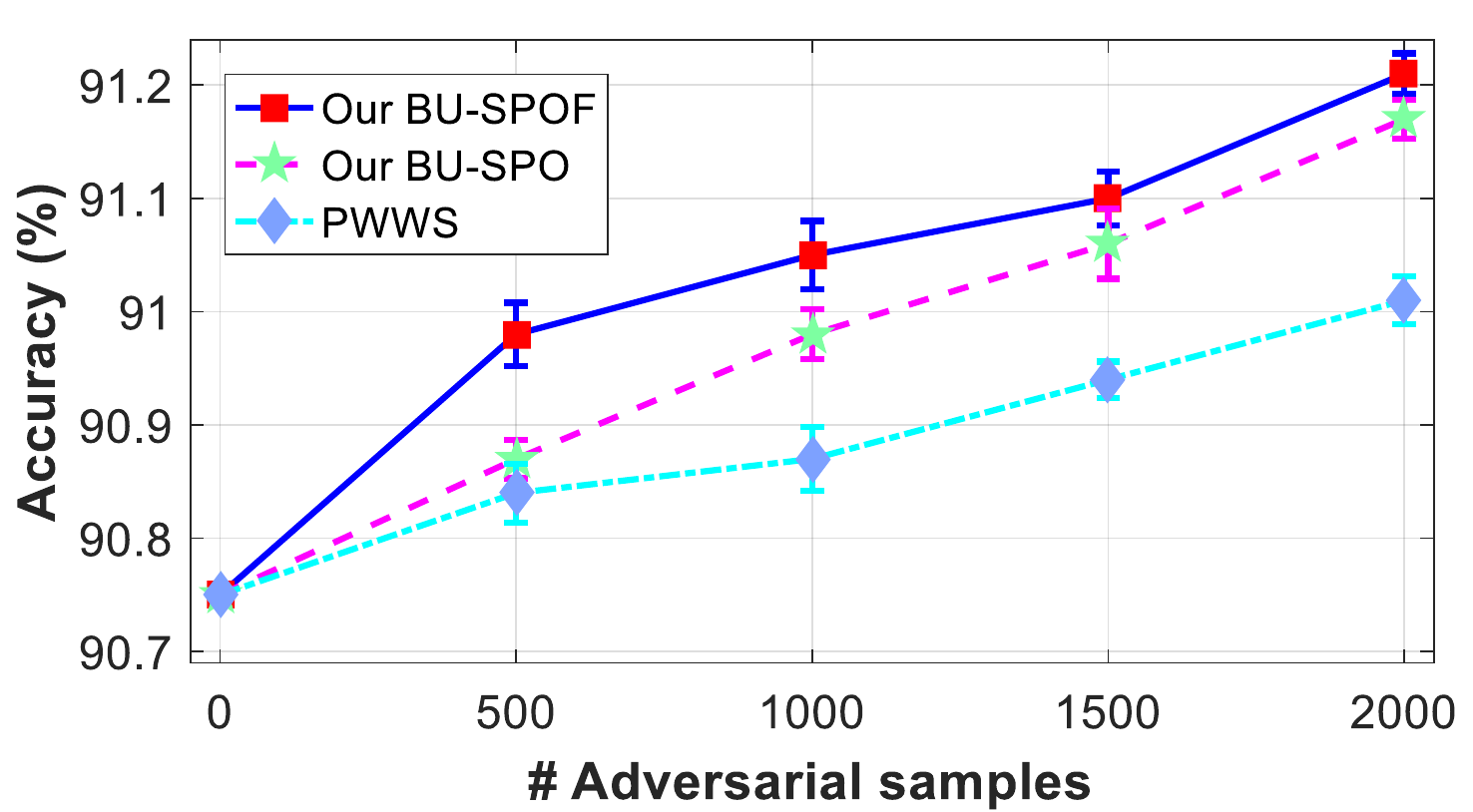}}
\caption{Adversarial retraining results. The higher the accuracy, the more robust of the model after retraining.}
\label{fig-AdversarialTraining}
\end{figure}

Adversarial retraining is an effective way to improve the model's robustness by joining the adversarial examples to the training set. In this experiment, we randomly select \{500, 1000, 1500, 2000\} AG's New training samples to generate adversarial examples. Then we append these crafted adversarial examples to the training set and retrain the Word CNN model. We evaluate if the adversarially retrained model becomes more robust by checking the classification accuracy on the test set. Fig. \ref{fig-AdversarialTraining} shows the five-run mean accuracy of Word CNN on the clean test set after adversarial training. From Fig. \ref{fig-AdversarialTraining} we obtain following two results: (1) the robustness of the retrained model is gradually improved when more adversarial examples are joined to the training set; (2) our BU-SPO and BU-SPOF methods generate more effective adversarial samples than PWWS in improving the model robustness. 

\begin{table}
\small
\centering
\caption{Targeted attack results on AG's News dataset.}
\begin{tabular}{p{1cm}m{0.8cm}<{\centering}m{0.9cm}<{\centering}m{1.35cm}<{\centering}m{0.75cm}<{\centering}m{1.35cm}<{\centering}}
\toprule
\multirow{2}[3]{*}{Model} & \multirow{2}[3]{*}{Target} & \multicolumn{2}{c}{ASR} &  \multicolumn{2}{c}{AWR} \\
\cmidrule(lr){3-4} \cmidrule(lr){5-6}
~                          & ~    & PWWS        & BU-SPOF              & PWWS       & BU-SPOF        \\
\midrule
\multirow{4}{*}{CNN}       & 0    & $87.21\%$   & \textbf{95.59\%}    & $4.78$     & \textbf{3.58} \\
~                          & 1    & $70.41\%$   & \textbf{91.14\%}    & $8.3$      & \textbf{5.47} \\
~                          & 2    & $73.66\%$   & \textbf{93.79\%}    & $7.52$     & \textbf{4.38} \\
~                          & 3    & $82.58\%$   & \textbf{93.41\%}    & $6.08$     & \textbf{4.41} \\
\midrule
\multirow{4}{*}{Ch-CNN}  & 0    & $32.61\%$   & \textbf{65.99\%}    & $11.56$    & \textbf{9.11} \\
~                          & 1    & $36.88\%$   & \textbf{66.18\%}    & $10.91$    & \textbf{8.79} \\
~                          & 2    & $46.58\%$   & \textbf{81.03\%}    & $9.11$     & \textbf{5.49} \\
~                          & 3    & $81.76\%$   & \textbf{96.05\%}    & $4.53$     & \textbf{2.81} \\
\midrule
\multirow{4}{*}{LSTM}      & 0    & $77.92\%$   & \textbf{90.79\%}    & $7.09$     & \textbf{5.21} \\
~                          & 1    & $72.27\%$   & \textbf{91.28\%}    & $7.88$     & \textbf{5.37} \\
~                          & 2    & $81.31\%$   & \textbf{90.99\%}    & $5.58$     & \textbf{4.64} \\
~                          & 3    & $79.50\%$   & \textbf{88.83\%}    & $6.74$     & \textbf{5.11} \\
\bottomrule
\end{tabular}
\label{Tab_targeted}
\end{table}

\subsection{Targeted Attack Evaluations}
\label{targeted}
Targeted attack is usually regarded as a more dangerous attack strategy, as it can arbitrarily mislead the victim model to misclassify any lable to a pre-specified label \cite{carlini2018audio}. In this section, we conduct the targeted attack experiments on AG's News dataset by attacking Word-CNN, Ch-CNN and Word-LSTM models. For each model, we attack 1000 legitimate samples to the four target labels: 0 (World), 1 (Sports), 2 (Business) and 3 (Sci/Tech). Table~\ref{Tab_targeted} shows the experimental results. From Table~\ref{Tab_targeted} we can see that our BU-SPOF attains a much higher ASR than PWWS for all target labels and victim models, especially for the Ch-CNN model. Besides, our BU-SPOF replaces less words than PWWS. This illustrates that our method is more powerful for both targeted attack and untargeted attack.

\section{Conclusions}
\label{conclusion}
In this paper, we have proposed a novel Bigram and Unigram based Semantic Preservation Optimization (BU-SPO) algorithm for crafting natural language adversarial samples. Specifically, the BU-SPO exploits both unigram and bigram modifications to avoid breaking commonly used bigram phrases. Besides, the hybrid synonym-sememe candidate selection approach provides better candidate options to craft high quality adversarial examples. More importantly, we design an adaptive SPO algorithm to determine the word substitution priority, which is significant in reducing the perturbation cost. We also proposed to improve the SPO with a semantic filter (BU-SPOF) to further enhance the semantic preservation performance. Extensive experimental results exhibit that our BU-SPO and BU-SPOF metohds achieve high attack success rates (ASR) and high semantic similarity with low numbers of word modifications. Besides, the proposed BU-SPOF also show its superiority on transfer attack, adversarial retraining, and targeted attack. In future, research on defense methods via using a n-gram strategy where n$>$2 will be a promising work direction.

\ifCLASSOPTIONcaptionsoff
  \newpage
\fi



%
\bibliographystyle{IEEEtran}
\bibliography{reference}

\begin{IEEEbiography}[{\includegraphics[width=1in,height=1.25in,clip,keepaspectratio]{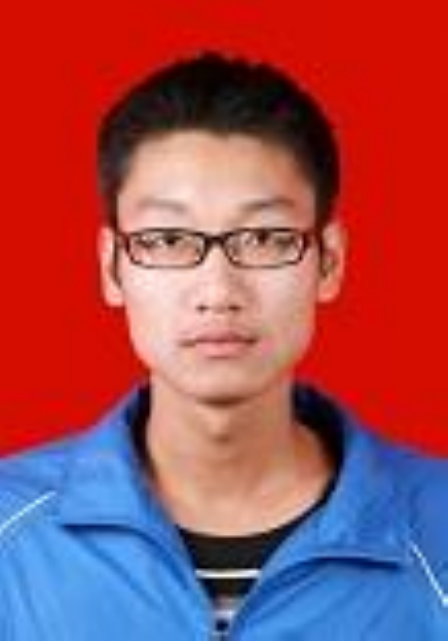}}]{Xinghao Yang}
received the B.Eng. degree in electronic information engineering and M.Eng. degree in information and communication engineering from the China University of Petroleum (East China), Qingdao, China, in 2015 and 2018, respectively. Currently, he is a PhD student in School of Computer Science, University of Technology Sydney, Australia. His research interests include multi-view learning and adversarial machine learning with publications on AAAI, IJCAI, TKDE, Information Fusion and Information Sciences.
\end{IEEEbiography}

\begin{IEEEbiography}[{\includegraphics[width=1in,height=1.25in,clip,keepaspectratio]{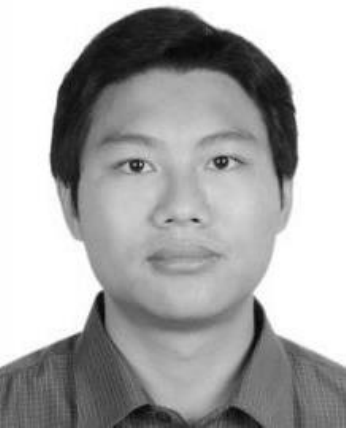}}]{Weifeng Liu}
(M'12-SM'17) received the double B.S.degrees in automation and business administration and the Ph.D. degree in pattern recognition and intelligent systems from the University of Science and Technology of China, Hefei, China, in 2002 and 2007, respectively. He is currently a Full Professor with the College of Information and Control Engineering, China University of Petroleum, Qingdao, China. He has authored or co-authored a dozen papers in top journals and prestigious conferences, including four Essential Science Indicators (ESI) highly cited papers and two ESI hot papers. His research interests include computer vision, pattern recognition, and machine learning. Prof. Liu serves as an Associate Editor for the Neural Processing Letters, the Co-Chair for the IEEE SMC Technical Committee on Cognitive Computing, and a Guest Editor for the special issue of the Signal Processing, the IET Computer Vision, the Neurocomputing, and the Remote Sensing. 
\end{IEEEbiography}

\begin{IEEEbiography}[{\includegraphics[width=1in,height=1.25in,clip,keepaspectratio]{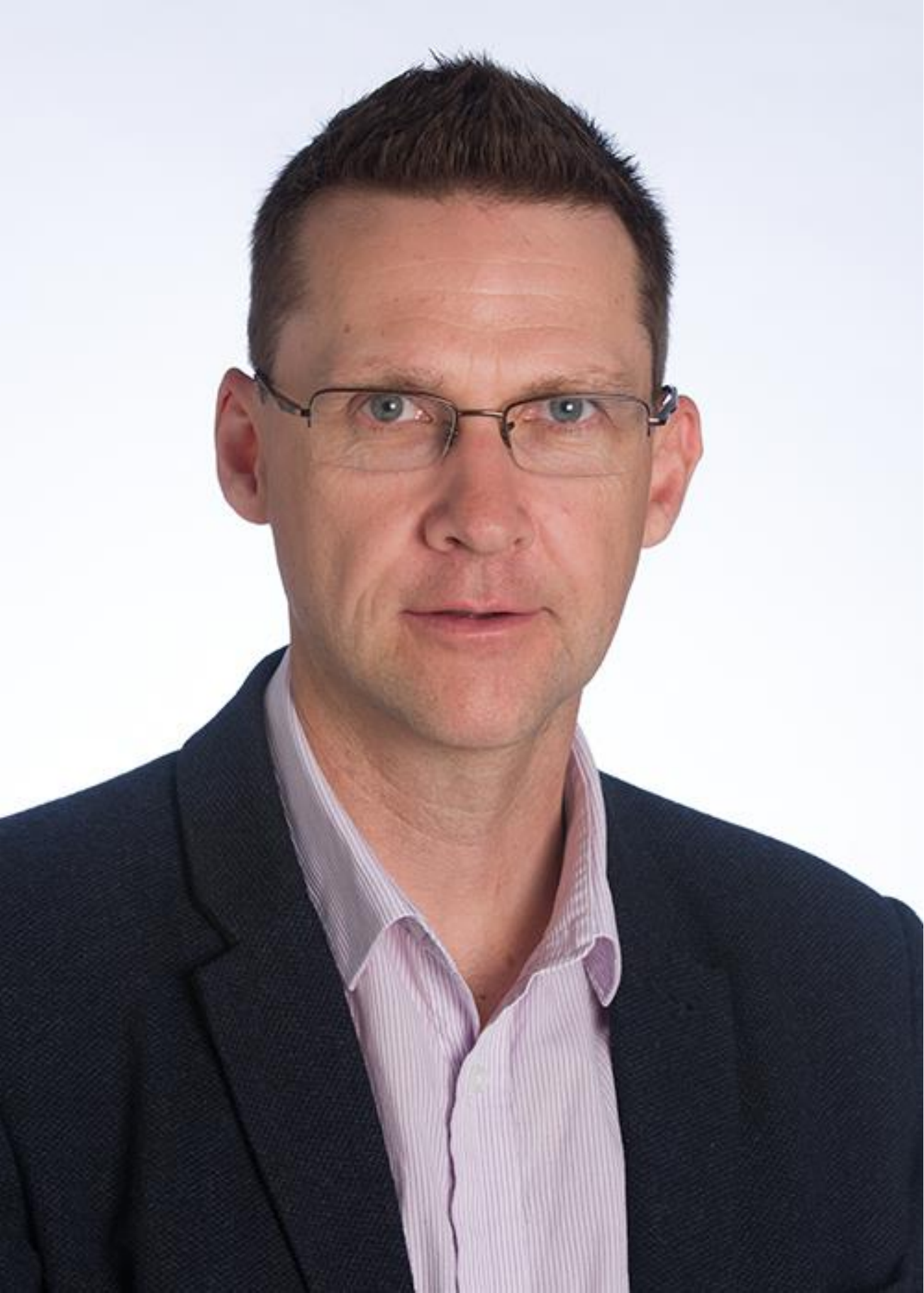}}]{James Bailey} received the PhD degree from the University of Melbourne, in 1998. He is a Professor in the School of Computing and Information Systems at the University of Melbourne. He was an Australian Research Council Future Fellow from 2012–2015. His research interests are in the area of data mining and machine learning, particularly perturbation analysis, clustering, correlation assessment and anomaly detection and explanation. His research has been translated to systems in the area of health, partnering with both hospitals (real time medical emergency prediction for patients) and industry (cognitive systems for immersive simulation training). He has received the best paper award at conferences such as IEEE ICDM, PAKDD and SIAM SDM. He was co-PC Chair of PAKDD 2016 and co-General Chair of ACM CIKM 2015. He is a member of several Editorial Boards, including ACM Transactions on Data Science, IEEE Transactions on Big Data, and Knowledge and Information Systems.
\end{IEEEbiography}

\begin{IEEEbiography}[{\includegraphics[width=1in,height=1.25in,clip,keepaspectratio]{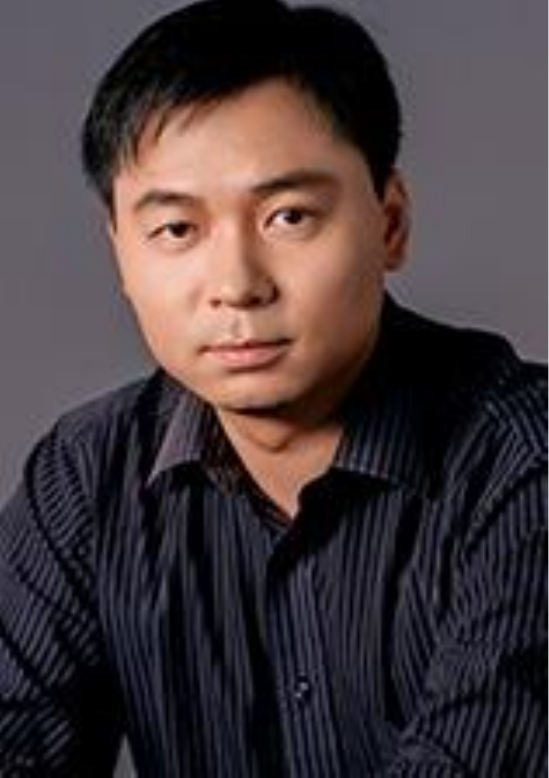}}]{Dacheng Tao}
(Fellow,  IEEE) is currently  the President  of  the  JD  Explore  Academy  and  a  Senior Vice   President   of   JD.com,   and   an   Advisor   and Chief  Scientist  of  the  Digital  Science  Institute  in The  University  of  Sydney.  He  mainly  applies  statistics  and mathematics  to artificial  intelligence  and data  science,  and  his  research  is  detailed  in  one monograph and over 200 publications  in prestigious journals and proceedings at leading conferences. 

Dacheng received the 2015/2020 Australian Eureka Prize, the 2018  IEEE  ICDM  Research  Contributions  Award, and  the  2021  IEEE  Computer  Society  McCluskey  Technical  Achievement Award. He is a fellow  of the Australian Academy of Science, AAAS, ACM, and IEEE.
\end{IEEEbiography}

\begin{IEEEbiography}[{\includegraphics[width=1in,height=1.25in,clip,keepaspectratio]{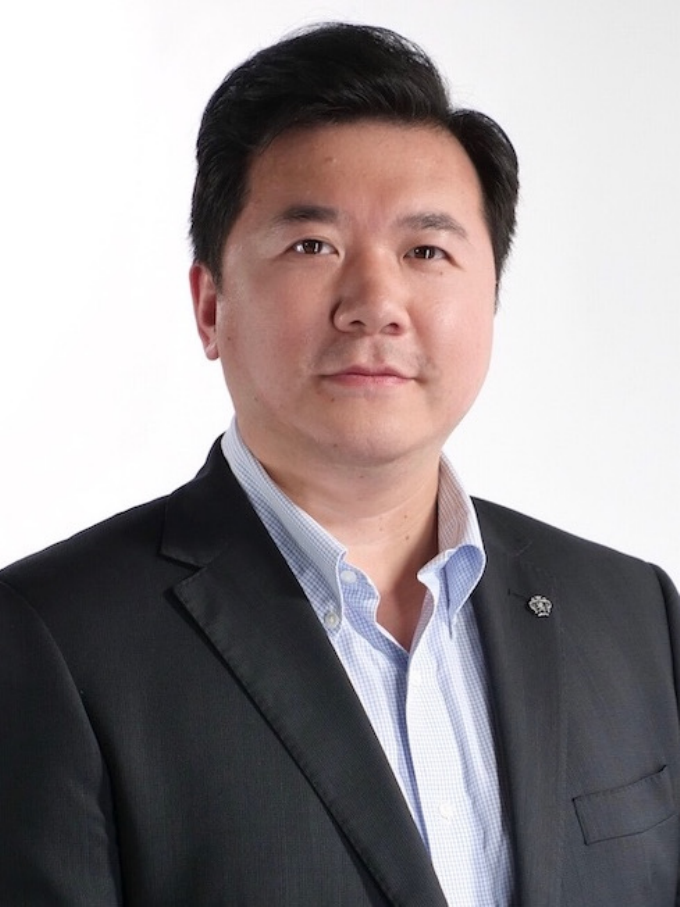}}]{Wei Liu} (M’15-SM’20) received the PhD degree in machine learning research from the University of Sydney in 2011. He is currently an Associate Professor and the Director of Future Intelligence Lab at the School of Computer Science, Faculty of Engineering and Information Technology, the University of Technology Sydney (UTS). Before joining UTS, he was a Research Fellow at the University of Melbourne and then a Machine Learning Researcher at NICTA. He works in the areas of machine learning and has published more than 90 papers in tensor factorization, adversarial learning, multimodal machine learning, graph mining, causal inference, and anomaly detection. 

Wei was a finalist of the 2017 NSW Premier's Prizes for NSW Early Career Researcher award. He has won three Best Paper Awards and one Most Influential Paper Award.
\end{IEEEbiography}

\end{document}